 \documentclass[authoryear,preprint,review,12pt]{elsarticle}

\usepackage{amssymb}

\journal{Computer Vision and Image Understanding}

\begin{document}

\begin{frontmatter}



\title{Unrepresentative video data: A review and evaluation }

\author{Georgios Mastorakis}
\ead{g.mastorakis@ieee.org}
\address{CogniLabs Ltd \& Kingston University London}

\begin{abstract}
It is well known that the quality and quantity of training data are significant factors which affect the development and performance of machine intelligence algorithms. Without representative data, neither scientists nor algorithms would be able to accurately capture the visual details of objects, actions or scenes. An evaluation methodology which filters data quality does not yet exist, and currently, the validation of the data depends solely on human factor. This study reviews several public datasets and discusses their limitations and issues regarding quality, feasibility, adaptation and availability of training data. A simple approach to evaluate (i.e. automatically ``clean" samples) training data is proposed with the use of real events recorded on the YouTube platform. This study focuses on action recognition data and particularly on human fall detection datasets. However, the limitations described in this paper apply in virtually all datasets.

\end{abstract}

\begin{keyword}
training data \sep machine learning \sep deep learning \sep datasets \sep fall detection \sep computer vision \sep data cleaning \sep action recognition


\end{keyword}

\end{frontmatter}


\section{Introduction}

Researchers in computer vision and data sciences require a significant amount of data to develop and validate their algorithms. There are several characteristics that define a good dataset to cover the range of relevant objects, actions or scenarios. A good dataset should include a sufficient number of examples to represent the variability of actions, human subjects, scene and light conditions, environmental changes and more. It should also provide annotated ground-truth of these objects, actions, etc. An example of how these datasets are acquired is the recording of videos of real-world scenes by CCTV cameras. The events shown in these datasets are in most cases accurate and of representative quality of a real-life event.

In this paper, the focus will be on a particular set of data which due to its complications in collecting, it is a good example of demonstrating the issues of training data. This set of data involves humans acting fall events. Ideally, one scenario for capturing such events would be to use cameras or other sensors in hospital wards, care homes, assisted living accommodation, older people's homes, rehabilitation centres, inpatient wards etc. Several of these data recording centres would be located around the globe in order to capture the human physiological characteristics variated by height, weight etc. Continuous recordings - day and night - of data over several months or years would have captured a significant number of falls as well as other activities of daily life (ADLs) with some having a similar motion pattern to a fall, such as lying down. The recordings would be in a format that protects personal privacy and allows public access and redistribution for scientific purposes. Unfortunately, the above scenario is imaginary as such recordings - even if they exist - are confidential, limited in variability and ethically unsuitable, as discussed later in this study. 

Therefore, to overcome the issue of data availability, researchers have implemented their versions of fall events - acted by volunteers aiming to bridge the gap between real and human-simulated falls. These datasets are discussed in this study in an attempt to show the limitations and difficulties in acquiring and using them. Recorded fall data acted by people is not as representative in relation to falling behaviour as when compared with datasets of other types of actions. In other words, an enacted walking behaviour is likely to be representative as it involves usual daily activities while enacting a fall is an artificial action subject to calculated behaviour. Given this reason, there is a scarcity of realistic fall samples due to hesitation \cite{gmiet2016} and the risk of injury performing a fall event. It is also notable, that usually only a small number of actors participate in these datasets for the above reason. These actors are mainly young and healthy while the vulnerable population (e.g. older or disabled people) is excluded from such samples due to ethical complications and risks and as a result human variability is limited.

Hesitation in falling is one of the measurable limitations/issues discussed above as it indicates how realistic a fall is, hence, based on the hesitation level, a researcher could compare enacted falls with real falling events. Apart from the review of fall datasets and the discussion of the limitations of data, this paper also proposes a methodology to evaluate and compare data from enacted falls to real data of people falling/fainting as a result of severe hyperventilation. This evaluation involves an automatic cleaning process that aims to select the most realistically performed samples and to filter out the ones in which hesitation in falling is high. This cleaning process is possibly the first described in the literature for data evaluation and it is an additional contribution of this study. 
 
A variety of different sensors such as RGB, RGB-D, accelerometers, gyroscopes and radars have been used to record fall events and a range of activities of daily life (ADLs) in order to record the negative samples of fall-like events, such as lying down. This study will focus on visual datasets and more particularly on the RGB and RGB-D datasets captured by 2D cameras and depth sensors.


Many relevant publicly available datasets have been created and are discussed in Sections \ref{public2Ddata} and \ref{publicDepthData}. Section \ref{dataLimitations} provides a discussion on the limitations of existing fall datasets considering their unsuitability in real environments due to occlusions and the limited representative nature of their demographics. Visuals from each dataset are included in order to provide a recognisable image of the data type. In order to highlight some of the issues of existing datasets and also to provide a data evaluation, the Section \ref{compCollapsing} presents a comparison between enacted and real fainting fall videos from YouTube. The next Section brings a more in depth understanding of fall events and their causes and characteristics.

\section{Causes and types of falls} \label{fallTypes}

Internal and external factors contribute to a fall, with the internal ones relating to the physical or mental state of the individual while the external ones relating to clothing, footwear \cite{kelsey2010footwear} and the environment. More specifically the physical factors, particularly for older people \cite{tinetti1986fall}, are mainly related to low/high blood pressure, brain atrophy \cite{yamada2013global}, low vision \cite{lord2001visual}, diabetes \cite{wallace2002incidence}, medication side-effects \cite{hartikainen2007medication}, muscle weakness \cite{moreland2004muscle}, vitamin deficiencies \cite{janssen2002vitamin}, injury of the lower limbs, gait irregularities \cite{weiss2013does} and balance issues. Mental conditions \cite{harlein2009fall, horikawa2005risk} may particularly affect cognition causing confusion, lack of attention, reduced sense of risk etc.

Falls have a direction according to the prior body motion or the centre of mass \cite{pai1997center}. These falls are directed towards the front, side or the backwards where the body stays relatively rigid and falls as a stick, or can have a vertical direction where the knees fold over and hit the ground first, while the rest of the body follows (i.e. collapsing fall). After such incidents, the person may remain unconscious on the ground or try to seek help. Trips and slips are considered as fall events which are caused by external factors such as elevated or slippery floor surfaces. These incidents may or may not conclude in an unconscious state, depending on the severity and location of the impact. Another group of falls is observed during sport events, where athletes fall unintentionally or intentionally in order to prevent an incident or pretend one. Other types of falls include the unexpected ones caused by an aggressive attack by another person or an animal.

This study considers mainly rigid and collapsing falls which lead to an unconscious state. These falling behaviours are found in the public datasets discussed in the next Sections.

\section{Camera (2D) datasets}
\label{public2Ddata}
Many early studies utilised RGB 2D cameras to record falls. These datasets are discussed in this section which refers to the early challenges of fall detection using this video data.

\subsection{Single camera LE2i dataset}

\begin{figure}
\centering
\includegraphics[width=0.9\textwidth]{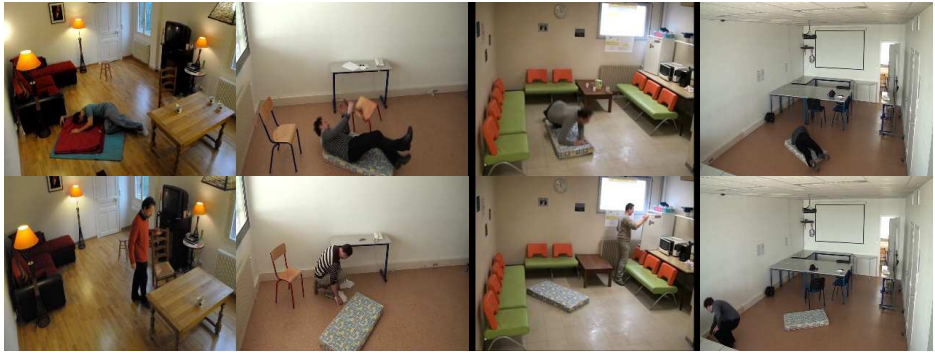}
\caption{Visuals from the LE2i dataset (\cite{charfi2013optimized}): fall events at top row, ADLs on lower row }
\label{fig:le2i}
\end{figure}

The LE2i dataset \cite{charfi2013optimized} contains 191 videos, 143 falls and 48 ADL of 9 subjects of unrecorded age, weight or height. The recordings are made in different types of rooms (i.e. at home, coffee room, office and lecture room) as seen in Figure \ref{fig:le2i} and according to the authors, this is done in order to evaluate the robustness of the method in different locations. Only one type of fall is shown: a rigid fall event with visible hesitation as actors place their hands towards the floor to minimise the impact as they fall. The data is captured using a single RGB camera and the video sequences are recorded using variable illumination.  The video also captures common visual difficulties, such as occlusions due to furniture or cluttered and textured background. The presence of occlusion is found in only 8 videos. However, the occlusion degree was minimal having small impact on the scenario, as seen in the first column of images in Figure \ref{fig:le2i}. 

\subsection{Multiple camera fall dataset}

The Multiple camera fall dataset \cite{auvinet2011fall} is one of the early attempts to record video data for the study of fall detection. One subject (i.e. of unknown age and other physical characteristics), performed 24 falls and 99 ADLs RGB videos. These actions included walking in different directions, housekeeping activities and actions with characteristics similar to falls (i.e. sitting down/standing up, crouching down, picking up an object from the floor). His falls involved a direction towards the front or backwards or failing to sit down properly or losing balance. The data collection used 8 cameras, mounted on the walls of a room perimetrically to record activity.

\begin{figure}
\centering
\includegraphics[width=0.9\textwidth]{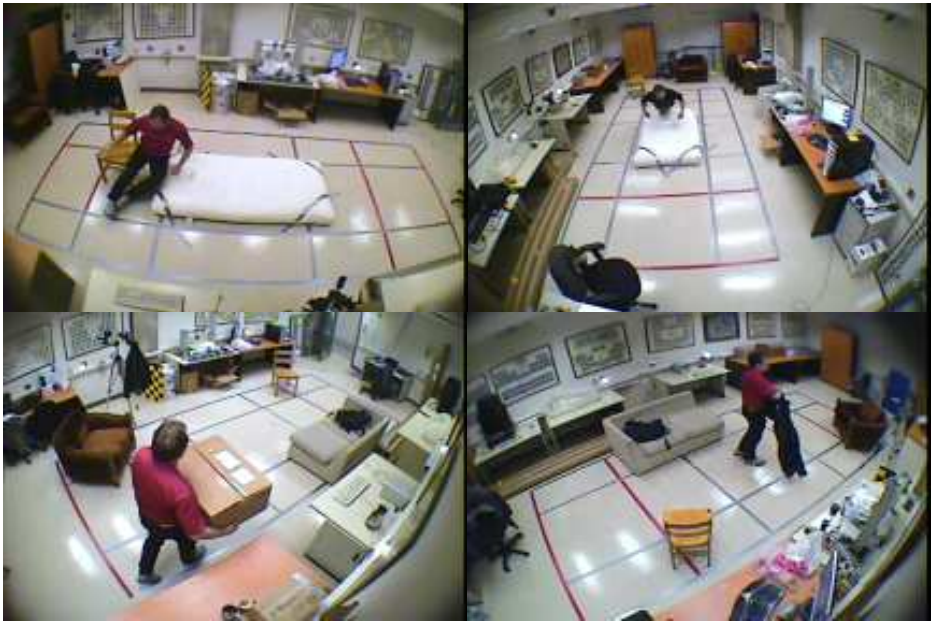}
\caption{Visuals from the Multiple cameras fall dataset (\cite{auvinet2011fall}): fall events at top row, ADLs on lower row}
\label{fig:Multiplecamerasfalldataset}
\end{figure}

Although there were objects that would potentially occlude the subject -- details of the size and location of these objects are only available visually from the images and they are not reported -- occlusion robustness is achieved from a non-occluded view of an event recorded from at least one of the cameras. Figure \ref{fig:Multiplecamerasfalldataset} shows several frames from this dataset of falls and ADL events.

\section{Public RGB-D datasets}
\label{publicDepthData}
More recently, RGB-D datasets have become publicly available for evaluating fall detection algorithms. The following, briefly summarises this composition.

\subsection{TST Fall Detection v2}

\begin{figure}
\centering
\includegraphics[width=0.9\textwidth]{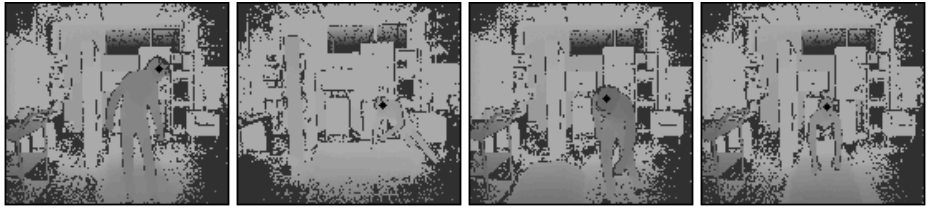}
\caption{Visuals from the TST Fall Detection v2 dataset (\cite{gasparrini2016proposal})}
\label{fig:tst2}
\end{figure}

The TST Fall Detection v2 \cite{gasparrini2016proposal} is an RGB-D dataset recorded using Microsoft Kinect v2 and two accelerometers placed on the wrist and waist of the subjects. This particular dataset was delivered by 11 subjects of unknown age, height or weight although the authors recorded some variation in height (1.62-1.97 m). Each subject performed 4 different ADLs (i.e. sitting down, walking and picking up an object from the floor, walking back and forth, lying down on a mattress) and 4 types of fall (i.e. falling forwards flat to the floor, falling to the side, falling backwards or sitting on the floor after a backwards fall). The falls concluded on the floor, however, the actions appeared rather rigid and staged.  Also, the format of the visual data makes this dataset difficult to process and as a result, this dataset is not used for evaluating other algorithms in the literature (i.e. it is a less preferable dataset). Figure \ref{fig:tst2} shows visuals from this dataset where fall or ADL events were performed facing the sensor.

\subsection{UR Fall Detection}

\begin{figure}
\centering
\includegraphics[width=0.9\textwidth]{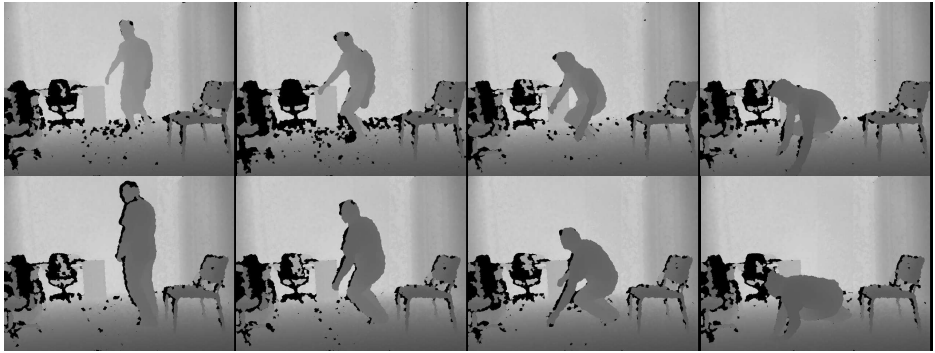}
\caption{Visuals from the UR Fall Detection dataset (\cite{kwolek2014human}) : a hesitated fall}
\label{fig:URFD}
\end{figure}

The UR Fall Detection \cite{kwolek2014human} is another dataset providing acceleration and video data (RGB and depth). It was collected using a two camera configuration -- one parallel to the floor and the other mounted on the ceiling. Annotations of other features, e.g. the bounding box around the person, were also provided. The dataset consisted of two types of falls: falls from a standing and a sitting position. This dataset is one of the most popular datasets that has been used by many researchers for their evaluation and comparison as it is rather straightforward to process (i.e. format is in PNG where pixel intensity denotes the correct depth). Nevertheless, this is a small dataset of 5 subjects performing only 15 falls from a standing position and another 15 from a sitting one. The subjects also appear to hesitate when performing a fall as seen in Figure \ref{fig:URFD} where falls conclude on the floor without using cushions or a mattress. On this video it is noticeable that the actor tries to reduce the impact by using his arms.

\subsection{SDUFall}

\begin{figure}
\centering
\includegraphics[width=0.9\textwidth]{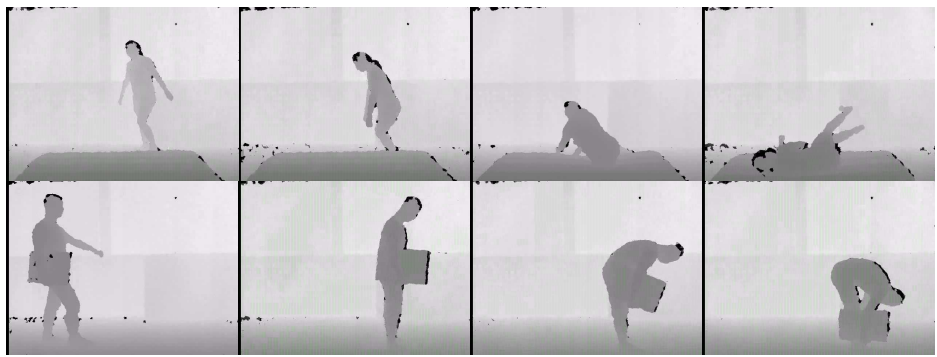}
\caption{Visuals from the SDUFall dataset (\cite{ma2014depth}) : fall events at top row, ADLs at lower row }
\label{fig:SDUFall}
\end{figure}

The SDUFall dataset \cite{ma2014depth} is one of the largest datasets comprising data captured from 20 people performing different types of falls (i.e. backwards, sideways) and 5 different ADLs (bending, squatting, sitting, lying and walking), with each subject repeating each action for 10 times. In each repetition of an ADL, the actors may or may not carry large objects, turn a light on or off, or change direction and position relative to the camera. This is another dataset where although there is a larger number of participants, the physical characteristics of each subject are not recorded. Many researchers have used this dataset as it consists of 200 fall samples recorded in depth signal, RGB and skeleton, distributed in .avi format and text files. Figure \ref{fig:SDUFall} shows visuals from this dataset where at the top row a fall occurs and at the lower row is picking up an object from the floor while holding a briefcase.

\subsection{University of Texas datasets}

Three datasets were collected at the University of Texas:

\subsubsection{The Falling Detection dataset} \cite{zhang2012viewpoint} dataset was collected in a laboratory-based simulated apartment set-up, with two Kinects mounted at opposite upper corners of the room. Six subjects performed 26 falls and several ADLs, such as picking up a coin from the floor, sitting down on the floor, tying shoelaces, lying down on a bed, opening a drawer which is close to the floor, jumping on to the floor and lying down on the floor. There were several ADLs samples recorded, however, the dataset provided only depth data and there was no information about the participants (i.e. physical caracteristics) or the sensors' exact location setup.

\subsubsection{The EDF dataset} \cite{zhang2015survey} extended the previous dataset in terms of data collection. The sensor setting remained the same in a simulated apartment where two Kinect sensors were installed to capture the events from two different directions, leading to a total of 320 sequences. Also, 100 sequences of 5 different ADLs, that could be associated with falls, were recorded i.e. ``picking up an object", ``sitting on the floor", ``lying down on the floor", ``tying shoelaces", and ``doing a plank exercise".

\subsubsection{The OCCU dataset} 
\begin{figure}
\centering
\includegraphics[width=0.9\textwidth]{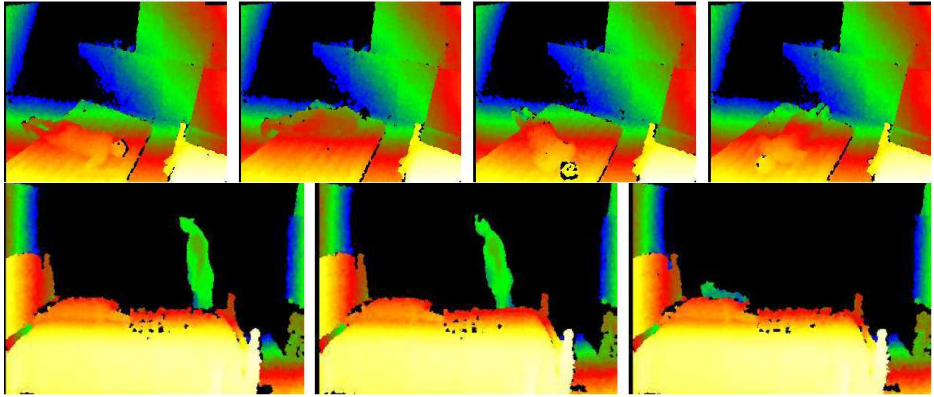}
\caption{Visuals from EDF (\cite{zhang2015survey}) and OCCU (\cite{zhang2014evaluating}) datasets. Top row: falls repetition over four angles, lower row: occluded fall behind a bed}
\label{fig:EDFOCCU}
\end{figure}

\cite{zhang2014evaluating}, as the EDF set, also used the same setting. The main feature of this dataset was the presence of occluded falls during which an object such as a bed, completely occluded the person at the end of the action. Five subjects simulated 12 falls, 6 from every viewpoint. Similarly to the EDF dataset, 80 sequences of actions that can be confused with falls were also provided. This is the only dataset where occlusions were introduced.
Figure \ref{fig:EDFOCCU} shows visuals from these datasets, where at the first row a fall occurs with different direction towards the sensor, while at the lower row an occluded fall occurs.

\subsection{ACT42 dataset}

\begin{figure}
\centering
\includegraphics[width=0.9\textwidth]{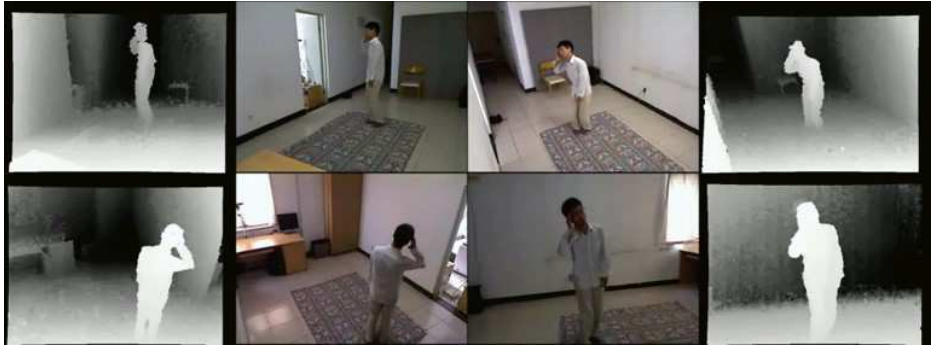}
\caption{Visuals from the ACT42 dataset (\cite{cheng2012human}). Events captured from four views in RGB-D}
\label{fig:ACT42}
\end{figure}

The ACT42 dataset \cite{cheng2012human} mainly focused on facilitating practical applications, such as a smart house or e-healthcare, and contained 14 daily activities, such as: drinking, making a phone call, moping the floor, picking up, putting on, reading a book, sitting down, sitting up, stumbling, taking off, throwing away, twisting open and wiping clean. Two categories of falls were considered, namely Collapsing (i.e. falling due to internal factors, i.e. heart attack, stroke etc.) and Stumbling (i.e. falling due to external obstacles). The dataset was captured by 4 Kinect sensors from different heights and view angles. This was one of the first datasets showing data of collapsing fall event videos. Nevertheless, in the majority of those videos, it is noticeable how subjects hesitate to fall in a vertical direction towards the ground. Also, data regarding the participants' physical characteristics was not available and the sensors' positions, although different in every capturing scenario, was not recorded (e.g. height of the sensor). Visuals from this datasets are seen in Figure \ref{fig:ACT42} where four cameras captured every event.

\subsection{Daily Living Activity Recognition}

\begin{figure}
\centering
\includegraphics[width=0.5\textwidth]{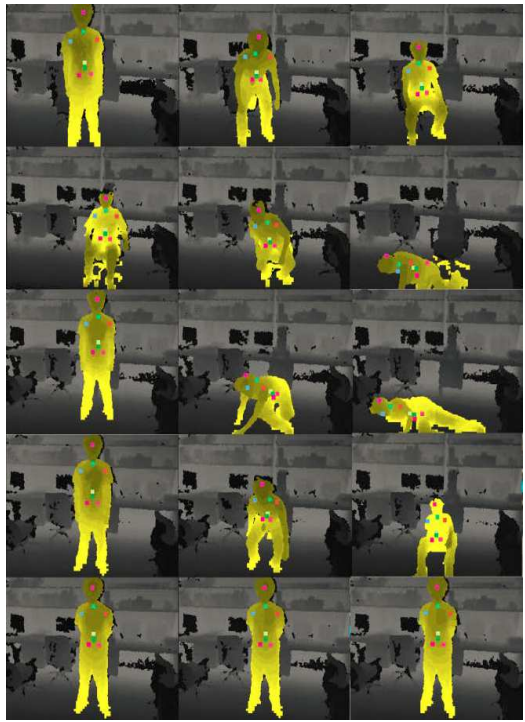}
\caption{Visuals from the Daily Living Activity Recognition dataset (\cite{zhang2012rgb}). Several ADLs and fall samples of RGB-D and skeleton}
\label{fig:DailyLivingActivityRecognition}
\end{figure}

The Daily Living Activity Recognition dataset \cite{zhang2012rgb} has data of subjects performing five activities related to falling events including standing, falling from standing, falling from sitting, sitting on a chair and sitting on the floor. It was captured using a Kinect sensor. RGB, depth and skeleton data were provided in this dataset in 150 different data files. Nevertheless, only 50 of these files are available for public retrieval. Subjects performed events in front of the sensor and without any occluded scenes. Other data was not recorded regarding the participants' characteristics or the sensor's location. Figure \ref{fig:DailyLivingActivityRecognition} shows several events from the dataset.

\subsection{NTU RGB+D Action Recognition Dataset}

\begin{figure}
\centering
\includegraphics[width=0.9\textwidth]{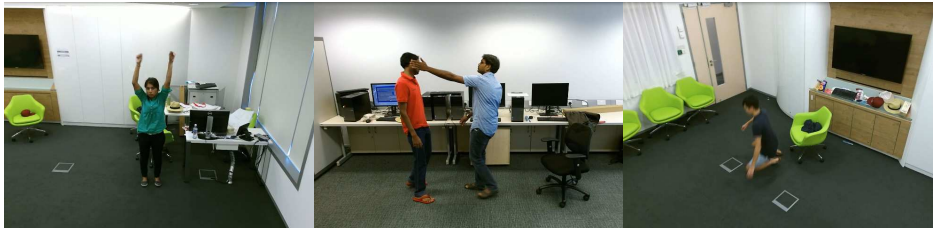}
\caption{Visuals from the NTU RGB+D Action Recognition dataset (\cite{shahroudy2016ntu}). Several ADLs and fall samples in RGB-D}
\label{fig:NTURGBD}
\end{figure}

This dataset \cite{shahroudy2016ntu} appears to have the most video samples of any set discussed in this study. This dataset was not prepared for fall detection studies as it contained only 40 fall events captured from different angles. The authors claim that there is a human variability on subjects such as age, height and weight, but this information was not made available. There were also videos where the fall event did not conclude to a resting place on the floor, but the subject stopped the fall by putting their arms forward. Falls appear to be conducted with minimum risk and hesitation is obvious. Also, the fall actions appear without occlusions from objects. Currently, at the time of writing this work, this dataset has not been used in a publication's bibliography for evaluating a fall detection algorithm. Figure \ref{fig:NTURGBD} shows visuals from the dataset where the first two images show ADL events and the last image shows a sideways fall.

\subsection{UWA3D Multiview Activity dataset} \label{multiview3D}

\begin{figure}
\centering
\includegraphics[width=0.9\textwidth]{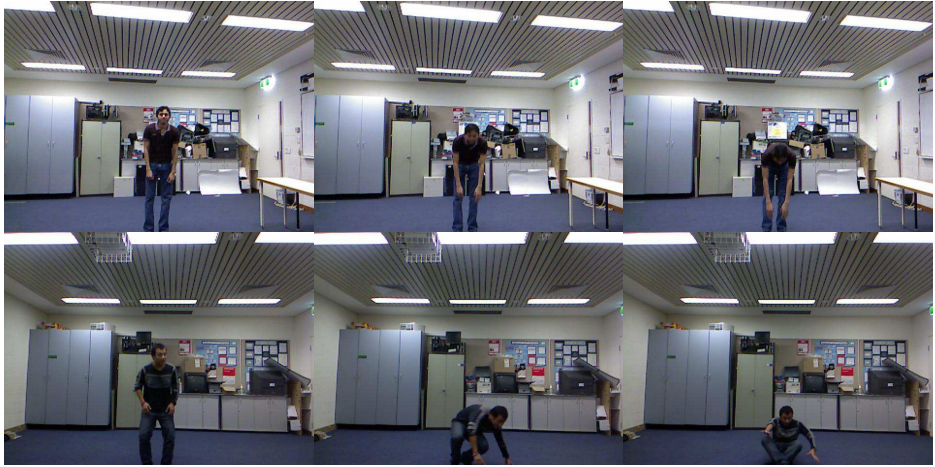}
\caption{Visuals from the UWA3D dataset (\cite{rahmani2014hopc}). Performance of a bending over (top row) and falling (low row)}
\label{fig:UWA3D}
\end{figure}

Dataset \cite{rahmani2014hopc} consists of 30 ADLs (i.e hand waving, one/two hands punching, sitting down, standing up, holding their chest, holding their head, touching their back, walking, turning around, drinking, bending, running, kicking, jumping, moping floor, sneezing, sitting down, squatting, two hands waving, vibrating, irregular walking, lying down, phone answering, jumping jack, picking up, putting down, dancing, and coughing) and a falling event, performed by 10 subjects. To achieve multi-view recording, five subjects performed 15 activities captured from four different views. Nevertheless, only the front view is available -- at the time of this study -- for retrieval. Subjects' physical characteristics were not discussed in the information related to participants. Visuals are shown in Figure \ref{fig:UWA3D} where at the first row a person is bending over, while at the lower row the person performs a collapse with noticeable hesitation.

\begin{table}[]
\centering
\scriptsize
\caption{RGB-D Fall datasets. R: RGB data, IR: infrared data, D: depth data, A: accelerometer data, S: Kinect skeleton data. The table shows the different fall event datasets of several sensor technologies. Noticeable is the number of fall events if compared with the ADLs as well as how small the fall number is in general}
\label{table:publicDatasets}
\begin{tabular}{c|c|c|c|c|c}
Dataset               & Subjects & Actions & \begin{tabular}[c]{@{}c@{}}Fall\\ Samples\end{tabular} & \begin{tabular}[c]{@{}c@{}}ADL\\ Samples\end{tabular} & \begin{tabular}[c]{@{}c@{}}Data\\ Type\end{tabular} \\
\hline
Multiple cameras~\cite{auvinet2011fall} & 1        & 9       & 24                                                     & 99                                                    &  R                                                 \\
LE2i~\cite{charfi2013optimized}                 & 9        & 7       & 143                                                    & 48                                                    &  R                                                 \\
TST v2~\cite{gasparrini2016proposal}                & 11       & 5       &                                                        &                                                       &  D, S, A                                             \\
UR~\cite{kwolek2014human}                    & 5        & 6       & 30                                                     & 40                                                    &  R, D, A                                           \\
SDUFall~\cite{ma2014depth}               & 20       & 6       & 200                                                    & 1000                                                  & R, D, S                                           \\
Fall Detection~\cite{zhang2012viewpoint}        & 6        & 8       & 26                                                     & 61                                                    & D                                                   \\
EDF~\cite{zhang2015survey}                  & 10       & 6       & 160                                                    & 50                                                    & D                                                   \\
OCCU~\cite{zhang2014evaluating}                  & 5        & 5       & 30                                                     & 80                                                    & D                                                   \\
ACT42~\cite{cheng2012human}                 & 24       & 14      & 48                                                     & 672                                                   & D, R                                              \\
Daily Living~\cite{zhang2012rgb}          & 5        & 5       & 10                                                     & 40                                                    & D, R, S                                           \\
NTU RGB+D~\cite{shahroudy2016ntu}             & 40       & 60      & 80                                                     & 4720                                                  & R, D, S, IR                                       \\
UWA3D~\cite{rahmani2014hopc}                 & 10       & 30      & 10                                                     & 290                                                   & R, D                                    \\         
\end{tabular}
\end{table}

\section{Limitations of existing datasets: A discussion}
\label{dataLimitations}

In general, as previously discussed, computer vision algorithms require a significant amount of data for training, which in this particular field, is sparse and of questionable quality in relation to how realistically the fall event is enacted. Table \ref{table:publicDatasets} summarises the samples found in public datasets and specifies the number of subjects and samples of each dataset. In the above-discussed datasets for action recognition, falls are a small class of experiments.

Genuine fall data is not readily available, particularly of vulnerable people as there are complications in collecting and distributing it. There are ethical reasons which prohibit older people and people with physical disabilities from participating in data collections that involve falls, due to the fragility of the body. The existing genuine data from actual scenes recorded in hospitals or assisted living homes is not available, mainly, due to reasons of privacy and ethical approval. 

As a result, researchers have implemented human-simulated falls in order to develop fall detection algorithms and fill the data availability gap. However, acting participants are asked to perform an event which in real life occurs spontaneously without calculated thought (e.g. a fall due to dizziness or stumbling on an obstacle). This implication makes the data collection a difficult task as the actors cannot perform realistically. A fall is not usually a daily event in real life and therefore it is difficult to replicate it genuinely for the purposes of data recording, unless studies use professional actors (i.e. stuntmen).

The following sections discuss in depth the issues with existing datasets and recording practices and provide the reader with an insight into their limitations.

\subsection{Age of participants}
All the datasets that have been discussed in this study provide limited data regarding the age of participants. In general, older people are not represented in any of the datasets - even if the actions are not fall related. Instead, generally young people perform the necessary fall event data recordings. These are mainly students and/or researchers from an academic institution that receive instructions to perform a fall as realistically as possible. In these circumstances, self-preservation takes over and the fall event is unrepresentative of a genuine fall, particularly if the aim is to acquire data representative of the vulnerable population (i.e. older people or people with physical disabilities). 

Particular emphasis is to be given to older people as the target group for fall detection applications. Older people are more prone to fainting due to various causes as seen in Section \ref{fallTypes}. Loss of balance can be due to muscle weakness or other medical conditions, such as reduced brain functionality, blood pressure issues, visual impairment, confusion and disorientation. Nevertheless, current datasets are used to train algorithms which aim to detect older people's falls without having representative data from this age group.


\subsection{Health of participants}
Participation in these datasets and performance of a fall requires the actors' physical condition and/or mental state to be assessed. If issues exist which conceal a risk, these participants will be excluded due to restrictions set by ethics committees, when human subjects are involved in data recording. Therefore, only the healthy individuals are allowed to participate while vulnerable population is excluded from the studies. Most of the existing work in fall detection aims to monitor and assist the vulnerable population in their daily life or even by preventing death, however, the investigation prohibits the recording of this data. 

\subsection{Types of falls}

The discussed datasets have samples from mainly one type of fall (i.e. rigid). The collapsing fall type is recorded in a few datasets but the falling behaviour is unrepresentative of a real collapsing event. One possible reason for avoiding the performance of such a fall, is the risk of injury, particularly in the knees \cite{KneeProblems}, which is a possibly why hesitation is significant. Section \ref{compCollapsing} tries to justify and assess the hesitation level in some of the collapsing falls.

\subsection{Size of datasets}
In most studies, the small number of actors performing fall events is not sufficient to represent the entire population. For example, one of the largest datasets \cite{shahroudy2016ntu} for fall detection consists of only 40 male and female subjects performing falls and other ADLs. Even if we could consider the number of participants as sufficient, it might not be sufficiently varied (see \ref{variability}). In reality, when compared with the number of recorded falls that occur every year in the US or the UK, it can be considered as small. Another factor for having small number of samples is the risk involved in the performance as previously discussed, which is a discouraging factor for perspective actors. The researchers themselves may also ask the actors to perform a small number of falls in an effort to prevent injuries. The small number of samples plays a significant role in the accuracy of a fall detection algorithm when applied to a real scenario, as the algorithm would have been trained on a small amount of data, resulting in a limited robustness.

\subsection{Variability of subjects}
\label{variability}
Variability in human physical characteristics associated with the height, weight, age, or gender, are factors which are not generally considered and they are not recorded by the developers of datasets. However, an older person has a different posture from a young one and a pregnant woman may walk differently from someone with a broken leg. Also, as seen in \cite{clauser1969weight}, men and women have a different centre of mass which affects the balance of the person. Any of these parameters can affect the falling behaviour. Collecting fall data from these groups may be difficult to impossible. However, it is important to be aware of the limited variability and that algorithms based on these datasets may have questionable performance when applied to broader demographics.

\subsection{Hesitation}
Subjects performing staged falls may have difficulty in acting realistically due to hesitation associated with the concern of having an injury. A \textit{hesitated fall} is defined in this study as a fall event where the person extends his/her arms to minimise the impact against the head or turns on the side to avoid an injury on the knees.

The risk of injury is an essential factor when permission is sought to conduct fall or other risky experiments; hence, the type of falls may be designed to follow a restricted protocol specified by regulations of health and safety or ethical considerations. Researchers have to warn the participants of any complication in the case of injury and may request disclaimers, especially if the falls are deployed in a real environment. As a result, data from such non-realistic recordings may have a negative impact on an algorithm's performance. 

\begin{figure}
\centering
\includegraphics[width=0.9\textwidth]{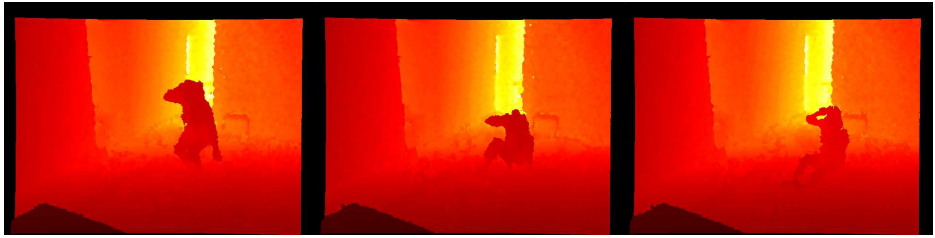}
\caption{Hesitation while collapsing. Subject uses an arm to balance, then expands her legs to sit on the floor (ACT42 dataset)}
\label{fig:collapsingHesitation}
\end{figure}

Visuals in Figure \ref{fig:collapsingHesitation} show hesitation due to self-preservation, resulting in an unrealistic fall example as the actor uses an arm to balance, then extends her legs to sit on the floor.

\subsection{Data quality and adaptation}
Other issues in using public datasets involve the encoding of data (i.e. video/image format) and how other researchers can adapt to this data format. In a few cases, depth data was compressed resulting in poor depth information and in other cases, the depth information was less reliable and as a result, further time was required to address the issues. For example, different depth sensors or OpenNI/Microsoft Kinect SDK versions delivered different video/image formats which was time-consuming to convert.

\subsection{Scene set-up}
Some of the discussed datasets including the RGB ones provided example events from actual home scenes. Although these scenes attempted to simulate a home environment, the rooms were sparsely furnished and unrealistically configured. Only a few occlusions were visible as most of the furniture was located near the walls.

\subsubsection{Occlusions}
\begin{figure}
\centering
\includegraphics[width=0.5\textwidth]{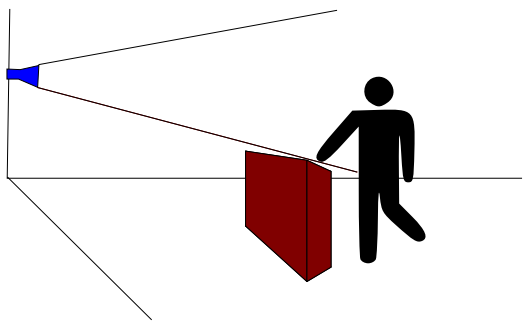}
\caption{Typical occluded scene. The camera view is partially blocked by the red box. Half the person is occluded.}
\label{fig:occlusionDiag}
\end{figure}

As noted from the analysis of the datasets in Sections \ref{public2Ddata} \ref{publicDepthData}, fall events appear fully visible in the video scenes without any scene occlusions. Datasets generally include fall events and ADL videos without the presence of other objects, unless this object is used (i.e. a chair, stool, bed) and occlusion scenarios are rarely represented. The lack of occlusion in most existing datasets is unrealistic for virtually all indoor (i.e. home) environments.  In a home scene we may get non-occluded views, but as people move around a cluttered environment, there may be frequent occasions during which they are partially occluded, to various degrees. Therefore, current algorithms are untested in the event of an occluded fall. Figure \ref{fig:occlusionDiag} illustrates an occlusion obstructing the view of a person.

Although many studies discuss the application of fall detection at home or in hospitals, occlusion is rarely mentioned; hence, methods are not evaluated to provide occlusion-robust solutions. A fall detector should rely on features that are visible and stable even in the event of an occluded home scene. The issue is that many fall detection algorithms may use one or more features that are more adversely affected by an occlusion on the ground plane (e.g. centre of mass). An approach to dealing with occlusions is to use several cameras, as seen in a few datasets, in order to maintain a continuous view of the scene, though this is still not guaranteed to eliminate the possibility of occlusion.

An attempt to evaluate current algorithms of datasets with occluded scenes is discussed in \cite{zhang2014evaluating} where authors have developed an occluded dataset and evaluated several state-of-the-art algorithms. Five subjects performed fall events which concluded with the fallen person completely occluded behind a bed. The level of occlusion that is estimated to be typically caused by a bed when the subject is at a standing position is relatively small to approximately 30\% of the person's height. However, the exact size/height of the bed was not provided and therefore the experiment did not accurately record the means of occlusion. Furthermore, the study failed to provide a proper evaluation of partially occluded ADLs such as walking, or picking up an object behind the bed, in order to measure the impact that an occlusion has on an non-fall event.

\subsubsection{Sensor location}
Only a few studies/datasets make a note of the sensor location. The position of the sensor plays a significant role as to where the best field-of-view is achieved in order to maintain a clear view of the home scene. This is unrelated to the minimisation of occlusions, because even if the sensor is located higher, occlusions may still occur. The sensor location in some cases plays a significant role in how the person appears; hence, an algorithm is designed to detect a fall using that type of data. See the example in \cite{kwolek2014human} where the depth sensor is located on the ceiling, pointing downwards. In other cases, the sensor is placed on a table, which seems unrealistic for a home scene. Also, by placing the sensor at a low height, the view is more prone to self-occlusions. In this scenario, a fall may start near the sensor and conclude on the floor in front of the sensor and possibly under the f-o-v of the sensor -- implying that the fall is outside the viewing window.

\section{Evaluation of enacted falls} \label{compCollapsing}

Using unrealistic and unrepresentative data significantly impacts on the performance of algorithms as discussed in the previous section. The development of a tool which assesses the data quality would immensely benefit both the development and evaluation of machine learning algorithms. Furthermore, with the development of deep learning \cite{lecun2015deep} the requirement of large datasets of representative and clean data has become a necessity. However, this process is quite time consuming and a rather intuitive process. In this study, a methodology to automatically assess the data quality is proposed. 

In this methodology, the level of realism of the actors' falling behaviour was assessed in order to filter the realistic and representative sample. A comparison between real and staged falls indicated which of the staged falls were representative. To investigate the level of realism, fall samples were taken from public video channels, such as YouTube. In recent years YouTube has become an increasingly useful source of video data -- although not all of the fall videos are of the right quality for processing. YouTube videos do not require ethical consideration and can be used freely. 

For this study, videos of young people hyperventilating until fainting, were used. Some of these videos show the realism and how serious a fall can be. YouTube hyperventilation videos were processed using a camera calibration software in order to measure the person's head vertical velocity $V_y$. Velocity was selected as a comparative feature as actors who hesitate generally try to slow down the fall by applying force to their knees or extend their arms to the ground in order to minimise the impact. This behaviour can affect the action and the velocity of the head would be different from the one in a real fall event.  

Noticeable hesitation was observed in collapsing fall videos found in  \cite{cheng2012human, shahroudy2016ntu, rahmani2014hopc} datasets. The subjects' head vertical velocity was measured in the above dataset. In order to measure the similarity between the velocity profiles of the real and enacted falls, the Hausdorff Distance (HD) was calculated, the validation of which is discussed in \cite{MASTORAKIS2018125}.

\begin{figure}
\centering
\includegraphics[width=0.99\textwidth]{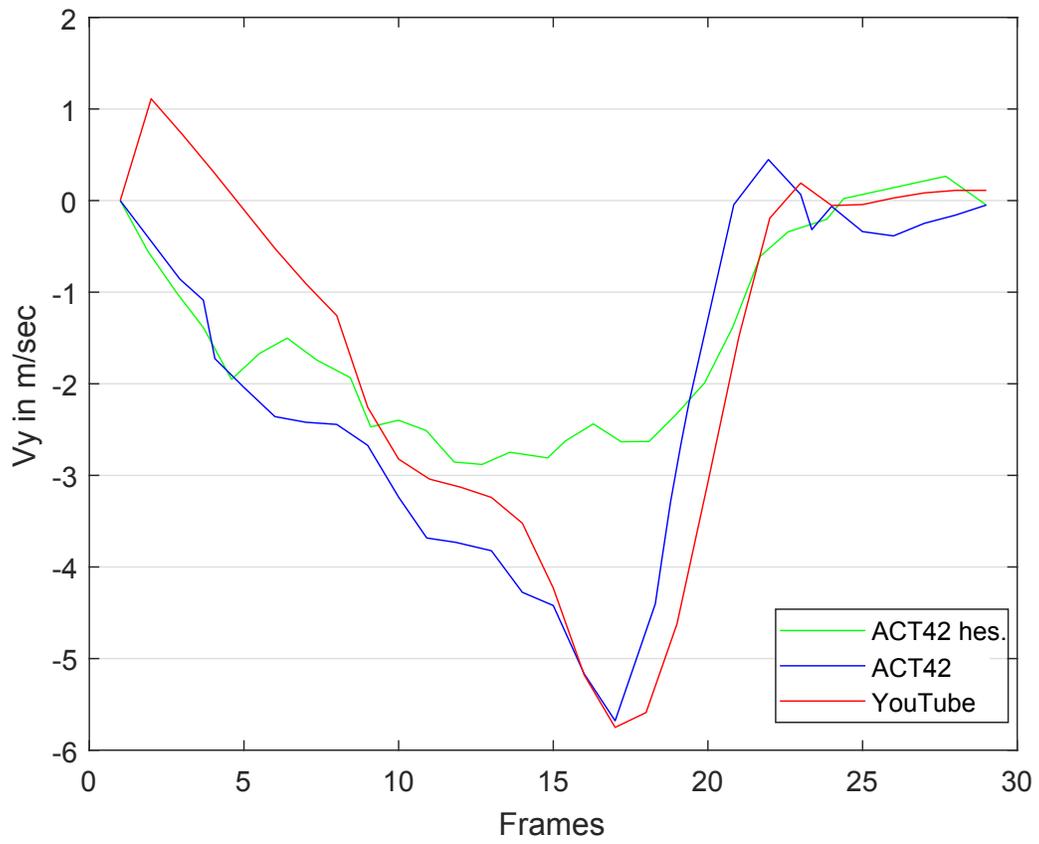}
\caption{Velocity profiles of collapsing falls. This Figure shows the velocity variation between a hesitated fall (ACT42 hes.) and an actual fall caused by hyperventilation }
\label{fig:collapsingComp2}
\end{figure}

\begin{figure}
\centering
\includegraphics[width=0.9\textwidth]{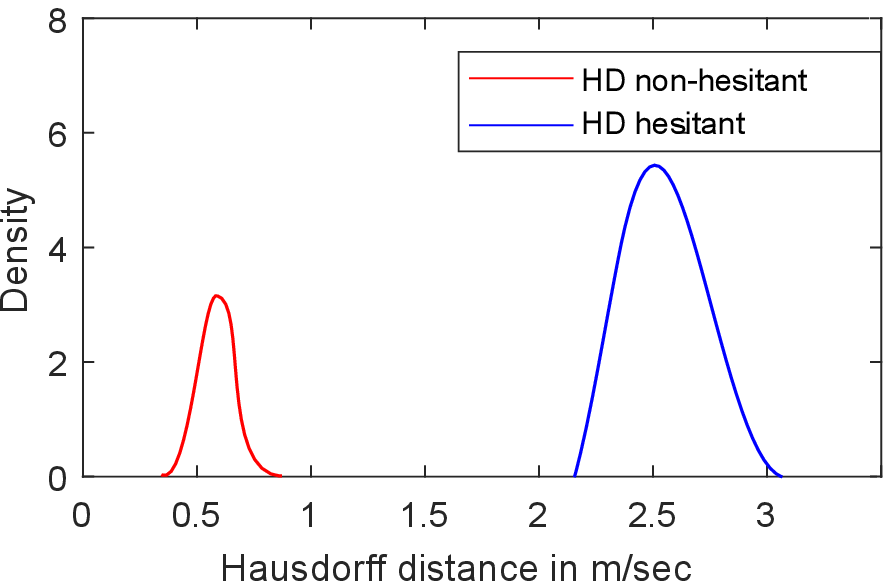}
\caption{Plot of a Gaussian pdf fitted to the distribution of Hausdorff distances: red curve denotes the HDs of YouTube to non-hesitant falls, while blue curve shows the HDs between YouTube and hesitant falls}
\label{fig:pdfHesitation}
\end{figure}

Using data samples from the ACT42 dataset, Figure \ref{fig:collapsingComp2} shows the velocity profiles of three different falls from the dataset (i.e. blue and green graph) and a real YouTube fall event (i.e. red graph). The Hausdorff distance between the YouTube and ACT42 hesitated profiles was 2.93 m/sec, while the HD of YouTube and a properly enacted collapsing example was 0.53 m/sec. To show the difference between the realistic and hesitated falls, a pdf was created as seen in Figure \ref{fig:pdfHesitation}. Two distributions are visible; the red graph presenting 7 examples of enacted realistic fall events being compared with YouTube ones and the blue graph presenting 41 samples of enacted collapsing falls which when compared with YouTube videos, they are classified as a similar to a non-fall event (see the evaluation of HD distance in \cite{MASTORAKIS2018125}). 

To summarise, only 7 examples from ACT42 were classified as realistic collapsing fall samples. This result indicates how actors hesitate in realistically performing a fall and hence, only a small sample is usable.

These results show how a valid collapsing fall can be evaluated and distinguished from a hesitated/unrealistic one processed by the ACT42 dataset. Samples with a small HD distance in velocity, when compared to a fall velocity profile, denote that their falling behaviour is more similar to a realistic fall, as evaluated in \cite{MASTORAKIS2018125}.







\section{Conclusion}

As discussed, current public datasets provide limited information regarding the actors' physical characteristics such as age, height, weight etc. which is insufficient in  order to accurately assess the impact of these characteristics on fall detection algorithms. Sometimes, the authors of these studies state the gender of the participants without specifying however, further information. Also, the number of fall events or of the participating actors is limited when compared to other action recognition datasets, due to the risks involved. In addition, these public datasets do not include data of the specific group for which they are usually intended i.e. the older people and people with disabilities and they have limited variability in terms of physical characteristics. Also, most of these datasets lack visual occlusions hence, there are no means to develop occlusion robust algorithms or assess their effectiveness. These issues may have serious implications in the deployment of fall detection systems in the wild. 

Hesitation is considered to be an important factor associated with unrealistic fall behaviour. This behaviour is observed in most types of falls, but it is more severe in collapsing events. Therefore, an evaluation of collapsing samples was undertaken in order to select and filter out the hesitated fall events. This study used comparative collapsing falls caused by hyperventilation recorded on YouTube. The procedure worked as a data cleaning mechanism that separated hesitated falls from realistic ones. This was done successfully by comparing the vertical velocity profiles of the head using the Hausdorff Distance.   

Machine learning would benefit from a further development of similar evaluation protocols utilising real samples from YouTube or other representative sources. These would develop new methodologies to assist in the ``data cleaning" process to select the representative samples for training and evaluation purposes. Therefore, a future direction of this evaluation would be the use of other actions -- where applicable/available -- or the use of physics-based simulation examples of human and object behaviour. This new pathway could positively improve the existing performance of algorithms.



\bibliographystyle{elsarticle-num-names} 
  \bibliography{arxidoBIB}

\begin{thebibliography}{29}
\expandafter\ifx\csname natexlab\endcsname\relax\def\natexlab#1{#1}\fi
\providecommand{\url}[1]{\texttt{#1}}
\providecommand{\href}[2]{#2}
\providecommand{\path}[1]{#1}
\providecommand{\DOIprefix}{doi:}
\providecommand{\ArXivprefix}{arXiv:}
\providecommand{\URLprefix}{URL: }
\providecommand{\Pubmedprefix}{pmid:}
\providecommand{\doi}[1]{\href{http://dx.doi.org/#1}{\path{#1}}}
\providecommand{\Pubmed}[1]{\href{pmid:#1}{\path{#1}}}
\providecommand{\bibinfo}[2]{#2}
\ifx\xfnm\relax \def\xfnm[#1]{\unskip,\space#1}\fi
\bibitem[{Mastorakis and Makris(2016)}]{gmiet2016}
\bibinfo{author}{G.~Mastorakis}, \bibinfo{author}{D.~Makris},
\newblock \bibinfo{title}{Fall detection: types, solutions, weaknesses. a
  review of computer vision based fall detection systems},
\newblock in: \bibinfo{booktitle}{IET Human Motion Analysis for Healthcare
  Applications}, \bibinfo{year}{2016}.
\bibitem[{LeCun et~al.(2015)LeCun, Bengio, and Hinton}]{lecun2015deep}
\bibinfo{author}{Y.~LeCun}, \bibinfo{author}{Y.~Bengio},
  \bibinfo{author}{G.~Hinton},
\newblock \bibinfo{title}{Deep learning},
\newblock \bibinfo{journal}{nature} \bibinfo{volume}{521}
  (\bibinfo{year}{2015}) \bibinfo{pages}{436}.
\bibitem[{Kelsey et~al.(2010)Kelsey, Procter-Gray, Nguyen, Li, Kiel, and
  Hannan}]{kelsey2010footwear}
\bibinfo{author}{J.~L. Kelsey}, \bibinfo{author}{E.~Procter-Gray},
  \bibinfo{author}{U.-S.~D. Nguyen}, \bibinfo{author}{W.~Li},
  \bibinfo{author}{D.~P. Kiel}, \bibinfo{author}{M.~T. Hannan},
\newblock \bibinfo{title}{Footwear and falls in the home among older
  individuals in the mobilize boston study},
\newblock \bibinfo{journal}{Footwear science} \bibinfo{volume}{2}
  (\bibinfo{year}{2010}) \bibinfo{pages}{123--129}.
\bibitem[{Tinetti et~al.(1986)Tinetti, Williams, and
  Mayewski}]{tinetti1986fall}
\bibinfo{author}{M.~E. Tinetti}, \bibinfo{author}{T.~F. Williams},
  \bibinfo{author}{R.~Mayewski},
\newblock \bibinfo{title}{Fall risk index for elderly patients based on number
  of chronic disabilities},
\newblock \bibinfo{journal}{The American journal of medicine}
  \bibinfo{volume}{80} (\bibinfo{year}{1986}) \bibinfo{pages}{429--434}.
\bibitem[{Yamada et~al.(2013)Yamada, Takechi, Mori, Aoyama, and
  Arai}]{yamada2013global}
\bibinfo{author}{M.~Yamada}, \bibinfo{author}{H.~Takechi},
  \bibinfo{author}{S.~Mori}, \bibinfo{author}{T.~Aoyama},
  \bibinfo{author}{H.~Arai},
\newblock \bibinfo{title}{Global brain atrophy is associated with physical
  performance and the risk of falls in older adults with cognitive impairment},
\newblock \bibinfo{journal}{Geriatrics \& gerontology international}
  \bibinfo{volume}{13} (\bibinfo{year}{2013}) \bibinfo{pages}{437--442}.
\bibitem[{Lord and Dayhew(2001)}]{lord2001visual}
\bibinfo{author}{S.~R. Lord}, \bibinfo{author}{J.~Dayhew},
\newblock \bibinfo{title}{Visual risk factors for falls in older people},
\newblock \bibinfo{journal}{Journal of the American Geriatrics Society}
  \bibinfo{volume}{49} (\bibinfo{year}{2001}) \bibinfo{pages}{508--515}.
\bibitem[{Wallace et~al.(2002)Wallace, Reiber, LeMaster, Smith, Sullivan,
  Hayes, and Vath}]{wallace2002incidence}
\bibinfo{author}{C.~Wallace}, \bibinfo{author}{G.~E. Reiber},
  \bibinfo{author}{J.~LeMaster}, \bibinfo{author}{D.~G. Smith},
  \bibinfo{author}{K.~Sullivan}, \bibinfo{author}{S.~Hayes},
  \bibinfo{author}{C.~Vath},
\newblock \bibinfo{title}{Incidence of falls, risk factors for falls, and
  fall-related fractures in individuals with diabetes and a prior foot ulcer},
\newblock \bibinfo{journal}{Diabetes care} \bibinfo{volume}{25}
  (\bibinfo{year}{2002}) \bibinfo{pages}{1983--1986}.
\bibitem[{Hartikainen et~al.(2007)Hartikainen, L{\"o}nnroos, and
  Louhivuori}]{hartikainen2007medication}
\bibinfo{author}{S.~Hartikainen}, \bibinfo{author}{E.~L{\"o}nnroos},
  \bibinfo{author}{K.~Louhivuori},
\newblock \bibinfo{title}{Medication as a risk factor for falls: critical
  systematic review},
\newblock \bibinfo{journal}{The Journals of Gerontology Series A: Biological
  Sciences and Medical Sciences} \bibinfo{volume}{62} (\bibinfo{year}{2007})
  \bibinfo{pages}{1172--1181}.
\bibitem[{Moreland et~al.(2004)Moreland, Richardson, Goldsmith, and
  Clase}]{moreland2004muscle}
\bibinfo{author}{J.~D. Moreland}, \bibinfo{author}{J.~A. Richardson},
  \bibinfo{author}{C.~H. Goldsmith}, \bibinfo{author}{C.~M. Clase},
\newblock \bibinfo{title}{Muscle weakness and falls in older adults: a
  systematic review and meta-analysis},
\newblock \bibinfo{journal}{Journal of the American Geriatrics Society}
  \bibinfo{volume}{52} (\bibinfo{year}{2004}) \bibinfo{pages}{1121--1129}.
\bibitem[{Janssen et~al.(2002)Janssen, Samson, and
  Verhaar}]{janssen2002vitamin}
\bibinfo{author}{H.~C. Janssen}, \bibinfo{author}{M.~M. Samson},
  \bibinfo{author}{H.~J. Verhaar},
\newblock \bibinfo{title}{Vitamin d deficiency, muscle function, and falls in
  elderly people},
\newblock \bibinfo{journal}{The American journal of clinical nutrition}
  \bibinfo{volume}{75} (\bibinfo{year}{2002}) \bibinfo{pages}{611--615}.
\bibitem[{Weiss et~al.(2013)Weiss, Brozgol, Dorfman, Herman, Shema, Giladi, and
  Hausdorff}]{weiss2013does}
\bibinfo{author}{A.~Weiss}, \bibinfo{author}{M.~Brozgol},
  \bibinfo{author}{M.~Dorfman}, \bibinfo{author}{T.~Herman},
  \bibinfo{author}{S.~Shema}, \bibinfo{author}{N.~Giladi},
  \bibinfo{author}{J.~M. Hausdorff},
\newblock \bibinfo{title}{Does the evaluation of gait quality during daily life
  provide insight into fall risk? a novel approach using 3-day accelerometer
  recordings},
\newblock \bibinfo{journal}{Neurorehabilitation and neural repair}
  \bibinfo{volume}{27} (\bibinfo{year}{2013}) \bibinfo{pages}{742--752}.
\bibitem[{H{\"a}rlein et~al.(2009)H{\"a}rlein, Dassen, Halfens, and
  Heinze}]{harlein2009fall}
\bibinfo{author}{J.~H{\"a}rlein}, \bibinfo{author}{T.~Dassen},
  \bibinfo{author}{R.~J. Halfens}, \bibinfo{author}{C.~Heinze},
\newblock \bibinfo{title}{Fall risk factors in older people with dementia or
  cognitive impairment: a systematic review},
\newblock \bibinfo{journal}{Journal of advanced nursing} \bibinfo{volume}{65}
  (\bibinfo{year}{2009}) \bibinfo{pages}{922--933}.
\bibitem[{Horikawa et~al.(2005)Horikawa, Matsui, Arai, SEKI, IWASAKI, and
  SASAKI}]{horikawa2005risk}
\bibinfo{author}{E.~Horikawa}, \bibinfo{author}{T.~Matsui},
  \bibinfo{author}{H.~Arai}, \bibinfo{author}{T.~SEKI},
  \bibinfo{author}{K.~IWASAKI}, \bibinfo{author}{H.~SASAKI},
\newblock \bibinfo{title}{Risk of falls in alzheimer’s disease: a prospective
  study},
\newblock \bibinfo{journal}{Internal Medicine} \bibinfo{volume}{44}
  (\bibinfo{year}{2005}) \bibinfo{pages}{717--721}.
\bibitem[{Pai and Patton(1997)}]{pai1997center}
\bibinfo{author}{Y.-C. Pai}, \bibinfo{author}{J.~Patton},
\newblock \bibinfo{title}{Center of mass velocity-position predictions for
  balance control},
\newblock \bibinfo{journal}{Journal of biomechanics} \bibinfo{volume}{30}
  (\bibinfo{year}{1997}) \bibinfo{pages}{347--354}.
\bibitem[{Charfi et~al.(2013)Charfi, Miteran, Dubois, Atri, and
  Tourki}]{charfi2013optimized}
\bibinfo{author}{I.~Charfi}, \bibinfo{author}{J.~Miteran},
  \bibinfo{author}{J.~Dubois}, \bibinfo{author}{M.~Atri},
  \bibinfo{author}{R.~Tourki},
\newblock \bibinfo{title}{Optimized spatio-temporal descriptors for real-time
  fall detection: comparison of support vector machine and adaboost-based
  classification},
\newblock \bibinfo{journal}{Journal of Electronic Imaging} \bibinfo{volume}{22}
  (\bibinfo{year}{2013}) \bibinfo{pages}{041106--041106}.
\bibitem[{Auvinet et~al.(2011)Auvinet, Multon, Saint-Arnaud, Rousseau, and
  Meunier}]{auvinet2011fall}
\bibinfo{author}{E.~Auvinet}, \bibinfo{author}{F.~Multon},
  \bibinfo{author}{A.~Saint-Arnaud}, \bibinfo{author}{J.~Rousseau},
  \bibinfo{author}{J.~Meunier},
\newblock \bibinfo{title}{Fall detection with multiple cameras: An
  occlusion-resistant method based on 3-d silhouette vertical distribution},
\newblock \bibinfo{journal}{IEEE Transactions on Information Technology in
  Biomedicine} \bibinfo{volume}{15} (\bibinfo{year}{2011})
  \bibinfo{pages}{290--300}.
\bibitem[{Gasparrini et~al.(2016)Gasparrini, Cippitelli, Gambi, Spinsante,
  W{\aa}hsl{\'e}n, Orhan, and Lindh}]{gasparrini2016proposal}
\bibinfo{author}{S.~Gasparrini}, \bibinfo{author}{E.~Cippitelli},
  \bibinfo{author}{E.~Gambi}, \bibinfo{author}{S.~Spinsante},
  \bibinfo{author}{J.~W{\aa}hsl{\'e}n}, \bibinfo{author}{I.~Orhan},
  \bibinfo{author}{T.~Lindh},
\newblock \bibinfo{title}{Proposal and experimental evaluation of fall
  detection solution based on wearable and depth data fusion},
\newblock in: \bibinfo{booktitle}{ICT innovations 2015},
  \bibinfo{publisher}{Springer}, \bibinfo{year}{2016}, pp.
  \bibinfo{pages}{99--108}.
\bibitem[{Kwolek and Kepski(2014)}]{kwolek2014human}
\bibinfo{author}{B.~Kwolek}, \bibinfo{author}{M.~Kepski},
\newblock \bibinfo{title}{Human fall detection on embedded platform using depth
  maps and wireless accelerometer},
\newblock \bibinfo{journal}{Computer methods and programs in biomedicine}
  \bibinfo{volume}{117} (\bibinfo{year}{2014}) \bibinfo{pages}{489--501}.
\bibitem[{Ma et~al.(2014)Ma, Wang, Xue, Zhou, Ji, and Li}]{ma2014depth}
\bibinfo{author}{X.~Ma}, \bibinfo{author}{H.~Wang}, \bibinfo{author}{B.~Xue},
  \bibinfo{author}{M.~Zhou}, \bibinfo{author}{B.~Ji}, \bibinfo{author}{Y.~Li},
\newblock \bibinfo{title}{Depth-based human fall detection via shape features
  and improved extreme learning machine},
\newblock \bibinfo{journal}{IEEE journal of biomedical and health informatics}
  \bibinfo{volume}{18} (\bibinfo{year}{2014}) \bibinfo{pages}{1915--1922}.
\bibitem[{Zhang et~al.(2012)Zhang, Liu, Metsis, and
  Athitsos}]{zhang2012viewpoint}
\bibinfo{author}{Z.~Zhang}, \bibinfo{author}{W.~Liu},
  \bibinfo{author}{V.~Metsis}, \bibinfo{author}{V.~Athitsos},
\newblock \bibinfo{title}{A viewpoint-independent statistical method for fall
  detection},
\newblock in: \bibinfo{booktitle}{Pattern Recognition (ICPR), 2012 21st
  International Conference on}, \bibinfo{organization}{IEEE},
  \bibinfo{year}{2012}, pp. \bibinfo{pages}{3626--3630}.
\bibitem[{Zhang et~al.(2015)Zhang, Conly, and Athitsos}]{zhang2015survey}
\bibinfo{author}{Z.~Zhang}, \bibinfo{author}{C.~Conly},
  \bibinfo{author}{V.~Athitsos},
\newblock \bibinfo{title}{A survey on vision-based fall detection},
\newblock in: \bibinfo{booktitle}{Proceedings of the 8th ACM International
  Conference on PErvasive Technologies Related to Assistive Environments},
  \bibinfo{organization}{ACM}, \bibinfo{year}{2015}, p.~\bibinfo{pages}{46}.
\bibitem[{Zhang et~al.(2014)Zhang, Conly, and Athitsos}]{zhang2014evaluating}
\bibinfo{author}{Z.~Zhang}, \bibinfo{author}{C.~Conly},
  \bibinfo{author}{V.~Athitsos},
\newblock \bibinfo{title}{Evaluating depth-based computer vision methods for
  fall detection under occlusions},
\newblock in: \bibinfo{booktitle}{International Symposium on Visual Computing},
  \bibinfo{organization}{Springer}, \bibinfo{year}{2014}, pp.
  \bibinfo{pages}{196--207}.
\bibitem[{Cheng et~al.(2012)Cheng, Qin, Ye, Huang, and Tian}]{cheng2012human}
\bibinfo{author}{Z.~Cheng}, \bibinfo{author}{L.~Qin}, \bibinfo{author}{Y.~Ye},
  \bibinfo{author}{Q.~Huang}, \bibinfo{author}{Q.~Tian},
\newblock \bibinfo{title}{Human daily action analysis with multi-view and
  color-depth data},
\newblock in: \bibinfo{booktitle}{Computer Vision--ECCV 2012. Workshops and
  Demonstrations}, \bibinfo{organization}{Springer}, \bibinfo{year}{2012}, pp.
  \bibinfo{pages}{52--61}.
\bibitem[{Zhang and Tian(2012)}]{zhang2012rgb}
\bibinfo{author}{C.~Zhang}, \bibinfo{author}{Y.~Tian},
\newblock \bibinfo{title}{Rgb-d camera-based daily living activity
  recognition},
\newblock \bibinfo{journal}{Journal of Computer Vision and Image Processing}
  \bibinfo{volume}{2} (\bibinfo{year}{2012}) \bibinfo{pages}{12}.
\bibitem[{Shahroudy et~al.(2016)Shahroudy, Liu, Ng, and
  Wang}]{shahroudy2016ntu}
\bibinfo{author}{A.~Shahroudy}, \bibinfo{author}{J.~Liu},
  \bibinfo{author}{T.-T. Ng}, \bibinfo{author}{G.~Wang},
\newblock \bibinfo{title}{Ntu rgb+ d: A large scale dataset for 3d human
  activity analysis},
\newblock in: \bibinfo{booktitle}{Proceedings of the IEEE Conference on
  Computer Vision and Pattern Recognition}, \bibinfo{year}{2016}, pp.
  \bibinfo{pages}{1010--1019}.
\bibitem[{Rahmani et~al.(2014)Rahmani, Mahmood, Huynh, and
  Mian}]{rahmani2014hopc}
\bibinfo{author}{H.~Rahmani}, \bibinfo{author}{A.~Mahmood},
  \bibinfo{author}{D.~Q. Huynh}, \bibinfo{author}{A.~Mian},
\newblock \bibinfo{title}{Hopc: Histogram of oriented principal components of
  3d pointclouds for action recognition},
\newblock in: \bibinfo{booktitle}{European Conference on Computer Vision},
  \bibinfo{organization}{Springer}, \bibinfo{year}{2014}, pp.
  \bibinfo{pages}{742--757}.
\bibitem[{WebMD(2018)}]{KneeProblems}
\bibinfo{author}{WebMD}, \bibinfo{title}{Knee problems and injuries},
  \bibinfo{howpublished}{https://www.webmd.com/pain-management/knee-pain/knee-problems-and-injuries-topic-overview},
  \bibinfo{year}{2018}. \bibinfo{note}{Accessed 11-Feb-2018}.
\bibitem[{Clauser et~al.(1969)Clauser, McConville, and
  Young}]{clauser1969weight}
\bibinfo{author}{C.~E. Clauser}, \bibinfo{author}{J.~T. McConville},
  \bibinfo{author}{J.~W. Young}, \bibinfo{title}{Weight, volume, and center of
  mass of segments of the human body}, \bibinfo{type}{Technical Report},
  ANTIOCH COLL YELLOW SPRINGS OH, \bibinfo{year}{1969}.
\bibitem[{Mastorakis et~al.(2018)Mastorakis, Ellis, and
  Makris}]{MASTORAKIS2018125}
\bibinfo{author}{G.~Mastorakis}, \bibinfo{author}{T.~Ellis},
  \bibinfo{author}{D.~Makris},
\newblock \bibinfo{title}{Fall detection without people: A simulation approach
  tackling video data scarcity},
\newblock \bibinfo{journal}{Expert Systems with Applications}
  \bibinfo{volume}{112} (\bibinfo{year}{2018}) \bibinfo{pages}{125 -- 137}.
  \DOIprefix\doi{https://doi.org/10.1016/j.eswa.2018.06.019}.

\end{thebibliography}




\end{document}